\documentclass[letterpaper, 10 pt, conference]{ieeeconf}  

\IEEEoverridecommandlockouts                              

\usepackage{amssymb}
\usepackage{amsmath}
\usepackage{graphicx}
\usepackage{colortbl}
\usepackage{xcolor}
\usepackage[utf8]{inputenc}
\usepackage[T1]{fontenc}
\usepackage{multirow} 
\usepackage{adjustbox}  %
\usepackage{booktabs}   %
\usepackage{graphicx}   
\usepackage{cite} 
\usepackage{array}
\usepackage{colortbl}
\usepackage{xcolor}
\usepackage{rotating}
\usepackage{lineno}
\usepackage{listings}
\usepackage{pmboxdraw} 
\usepackage{makecell}
\usepackage{tikz}
\usepackage{titlesec}
\usepackage{pgfplots}
\usepackage{pgfplotstable}
\usepackage[justification=justified]{caption}
\usepackage{bbding}
\usepackage{makecell}
\usepackage{multirow}
\usepackage{placeins}
\usepackage{wrapfig}
\usepackage{float}
\usepackage{pifont}

\title{\LARGE \bf
UnsOcc: 3D Semantic Occupancy Prediction in Unstructured Scene via Rendering Fusion
}

\author{Ye Wu$^{1*}$, Ruiqi Song$^{2,3*}$, Baiyong Ding$^{2,3}$, Nanxin Zeng$^{1}$, Junjie Cheng$^{1}$ and Yunfeng Ai$^{1,3\dag}$
\thanks{This work was supported by the Key Research and Development Program of Shaanxi Province under Grant 2024CY2-GJHX-49 and the Industry-University-Research Innovation Fund for Chinese Universities under Grant 2024HT023.}
\thanks{$^{1}$School of Artificial Intelligence, University of Chinese Academy of Sciences, Beijing 100049, China
        {\tt\small \{wuye23, zengnanxin24, chengjunjie25\}@mails.ucas.ac.cn, aiyunfeng@ucas.ac.cn}}
\thanks{$^{2}$Institute of Automation, Chinese Academy of Sciences, Beijing, China
        {\tt\small ruiqi.song@ia.ac.cn, knightdby@gmail.com}}
\thanks{$^{3}$Waytous Inc., Beijing, China}
\thanks{$^{\dag}$Corresponding author.}
\thanks{\textsuperscript{*}These authors contributed equally to this work.}
}

\begin{document}
\voffset=3pt
\maketitle

\thispagestyle{empty}
\pagestyle{empty}

\begin{abstract}
Unstructured scenes present unique challenges for autonomous driving, as irregular obstacles and sparse scene layouts undermine the effectiveness of traditional perception methods such as 3D object detection. 3D semantic occupancy prediction has emerged as a prominent focus due to its ability to provide dense spatial representations by assigning semantic labels to individual voxels in 3D space. However, directly applying 3D semantic occupancy prediction to unstructured scenes remains challenging because scene sparsity hinders effective cross-modal fusion and the more severe long-tail distribution in these scenarios further degrades prediction performance. To validate the effectiveness of our approach, we construct a dedicated dataset of unstructured scenes collected from open-pit mines. Based on this, we propose UnsOcc, a multi-modal 3D semantic occupancy prediction framework that improves robustness in unstructured environments. At its core, we introduce a rendering-based fusion module, RenderFusion, which enhances cross-modal feature alignment through bidirectional rendering supervision. Furthermore, we propose GSRefinement, a detail-aware auxiliary supervision method based on Gaussian Splatting that projects sparse 3D occupancy predictions into dense 2D semantic segmentation maps, enabling effective supervision for long-tail categories. Extensive experiments on both the open-pit mine dataset and the nuScenes dataset demonstrate that our method significantly outperforms existing state-of-the-art approaches.
\end{abstract}


\section{INTRODUCTION}

With the continuous advancement of autonomous driving technology, its application scenarios have progressively extended to more complex unstructured scenes, such as the surface mine. However, traditional perception methods, such as 3D object detection\cite{lang2019pointpillars,liu2023bevfusion}, often suffer performance degradation when confronted with irregular and long-tail obstacles, limiting their applicability in practical applications.
To address these challenges, 3D occupancy prediction has emerged as a promising alternative. Rather than relying on 3D bounding boxes, this method divides 3D space into voxel grids and assigns a semantic label to each voxel, giving an efficient 3D semantic representation for unstructured scenes. 

Currently, most 3D semantic occupancy prediction methods rely solely on image inputs\cite{cao2022monoscene, wei2023surroundocc, zhang2023occformer}. While images provide abundant texture and semantic information, they lack real-world scale and fail to provide absolute depth information. In contrast, LiDAR sensors provide 3D point clouds at centimeter level with precise distance measurements but suffer from relatively low spatial resolution and lack semantic information.

Therefore, fusing image and LiDAR modalities is essential to improve 3D perception performance, especially in complex unstructured scene. Although there have been several studies on multimodal fusion for 3D semantic occupancy prediction \cite{wang2023openoccupancy, ming2024occfusion}, effective multi-sensor fusion for unstructured scenes, characterized by sparse geometric features and weak semantic cues, remains underexplored \cite{chen2020novel}.
In addition, the long-tail distribution issue is further exacerbated in unstructured scenes.

\begin{figure}[t] 
\centering 
\includegraphics[width=0.48\textwidth]{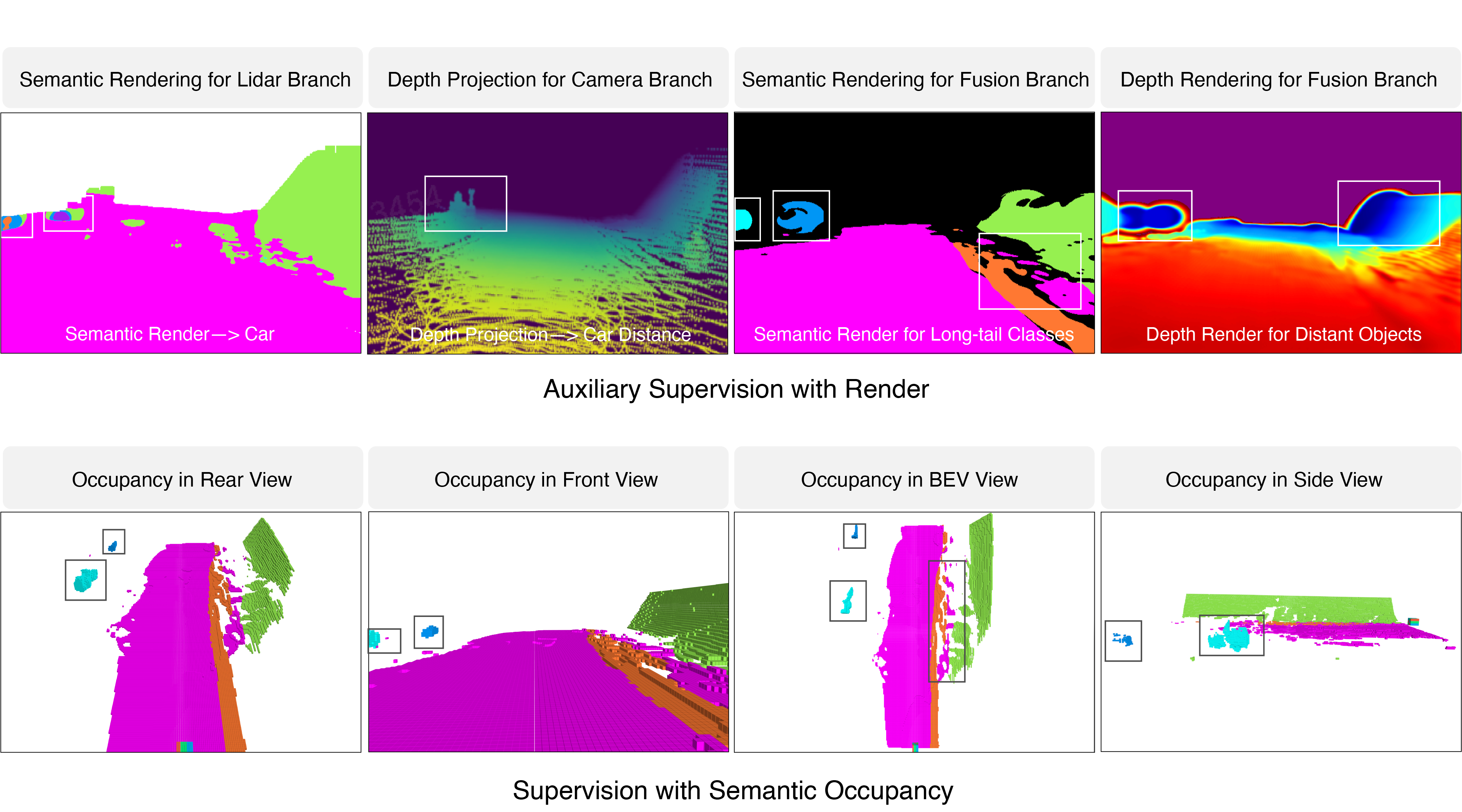}
\caption{
Semantic and Depth Rendering in Unstructured Scenes with 3D Gaussian Splatting. 
}
\vspace{-6mm}
\label{fig:vis}
\end{figure}
To tackle these challenges, we propose UnsOcc, a novel multi-modal fusion framework for 3D semantic occupancy prediction in unstructured scenes. It introduces RenderFusion, a render-based cross-modal alignment mechanism, and GSRefinement, a detail-aware 3D Gaussian Splatting auxiliary supervision module.
RenderFusion performs cross-branch rendering to align image and LiDAR features. The image branch uses a depth prediction network supervised by projected LiDAR points, while the LiDAR branch predicts semantic occupancy and renders it into 2D semantics via Gaussian Splatting, supervised by image-based segmentation. This dual-supervision enhances cross-modal feature alignment and fusion.
Furthermore, GSRefinement is used to address the severe sparsity of occupancy information in 3D voxel space. During the training phase, it not only predicts 3D semantic occupancy distributions but also projects them into 2D semantic segmentation maps for auxiliary supervision, with an emphasis on improving recognition of long-tailed classes. This strategy effectively improves the model’s ability to predict long-tailed classes. Figure \ref{fig:vis} illustrates a series of rendering results produced by our method.
To facilitate the deployment of autonomous driving in unstructured scenes, we developed a n open-pit mine dataset for 3D semantic occupancy prediction and conducted related experiments.

To summarize, our main contributions are as follows: 1) We propose a novel 3D semantic occupancy prediction framework that improves the robustness of prediction in unstructured scenes. 2) Bidirectional supervision for cross-modal feature alignment mechanism and detail-aware 3D Gaussian Splatting auxiliary supervision mechanism are proposed to enhance the capability of cross-modal fusion and long-tail class prediction in unstructured scene, respectively.
3) A dataset specifically designed for autonomous driving in unstructured scene has been constructed, and extensive experiments have been conducted.
\section{Related Work}
\subsection{3D Semantic Occupancy Prediction}
3D occupancy prediction aims to estimate the geometric and semantic occupancy of the surrounding space, which is crucial for autonomous driving and robotics. Early approaches to 3D occupancy prediction primarily relied on monocular depth estimation\cite{cao2022monoscene, miao2023occdepth, li2023voxformer}. MonoScene\cite{cao2022monoscene} and OccDepth\cite{miao2023occdepth} lifted 2D features into 3D voxel grids through monocular depth estimation. VoxFormer\cite{li2023voxformer} generates reliable queries by predicting image depth, enabling more accurate 3D scene understanding.While these approaches proved effective under limited camera views, they struggled with depth ambiguities, particularly in challenging scenarios. To overcome these limitations, panoramic or surround-view methods were subsequently proposed\cite{tian2023occ3d, tong2023scene, wei2023surroundocc}. Techniques like OccNet\cite{tong2023scene} and Occ3D\cite{wei2023surroundocc} leverage multiple camera perspectives to generate more accurate 3D occupancy maps, effectively mitigating depth ambiguities through cross-view information. However, these methods still rely on dense voxel grids, which incur significant computational costs. As a result, recent approaches have begun to explore alternative representations beyond voxels. \cite{shi2024occupancy} replaces dense voxel grids with point-based representations, which allow inputs of arbitrary scale and location, increasing modeling flexibility. TPVFormer\cite{huang2023tri} extends the BEV by adding two additional perpendicular planes to form the Tri-Perspective View (TPV), replacing voxel-based representations and achieving a balance between performance and computational complexity. GaussianFormer\cite{huang2024gaussianformer} utilizes sparse Gaussian primitives to model the scene, achieving accurate scene perception through the optimization of their properties. In this paper, we propose a novel alignment fusion scheme and projection-based supervision to enhance the performance of 3D occupancy prediction in unstructured scenes.
\subsection{Scene Representation with Rendering}
NeRF\cite{mildenhall2021nerf} introduced a neural rendering framework that models a scene as a continuous volumetric function, synthesizing novel views from sparse input images. It has inspired various scene reconstruction techniques\cite{zhang2020nerf++, barron2021mip}, with some adaptations for scene perception tasks\cite{zhang2023occnerf, pan2024renderocc}. However, NeRF-based approaches often incur high computational costs. 
To overcome these limitations, 3D Gaussian Splatting\cite{kerbl20233d} has been proposed as an efficient and explicit alternative. By using a set of anisotropic Gaussian primitives and employing splat rendering, 3D GS enables real-time rendering. Compared to NeRF, the explicit use of Gaussian primitives allows for a more direct and clear representation of geometric structure and occupancy information, while offering higher efficiency in rendering large-scale scenes. Due to these advantages, existing scene perception methods have also adopted 3D Gaussian Splatting\cite{huang2024gaussianformer, gan2024gaussianocc, jiang2024gausstr, chabot2025gaussianbev, chambon2025gaussrender}. GaussianFormer\cite{huang2024gaussianformer} adopts sparse 3D semantic Gaussian primitives to represent the scene. It iteratively optimizes the properties of the Gaussian primitives and employs an efficient Gaussian-to-voxel splatting method to generate 3D occupancy predictions. GaussRender\cite{chambon2025gaussrender} is not confined to projecting 3D voxels onto the camera viewpoint. It introduces the projection of 3D occupancy predictions and ground truth to arbitrary viewpoints for loss computation. Gaussianbev\cite{chabot2025gaussianbev} models the scene with a set of 3D Gaussian primitives and leverages 3D Gaussian splatting, in place of LSS method, to acquire BEV representation.

\subsection{Feature Fusion for 3D Scene Representation}

The image and LiDAR modalities exhibit significant complementarity in perception tasks. 
Therefore, fusing image and LiDAR data can overcome the limitations of each individual modality, offering a more comprehensive and robust scene perception capability.
A large number of perception solutions, including 3D object detection\cite{ob3, ob4} and semantic segmentation tasks\cite{seg1, seg2}, have already adopted fusion schemes that combine image and LiDAR data. In the context of 3D occupancy prediction tasks, there has also been research focusing on multimodal fusion\cite{ming2024occfusion, pan2024co, wang2023openoccupancy, wang2024occgen}. OccFusion\cite{ming2024occfusion} fuses the mapped image features and point cloud features by concatenating them along the channel dimension and applying 3D convolutions. Co-Occ\cite{pan2024co} introduces the K-Nearest Neighbor (KNN) algorithm to optimize the concatenation process, alleviating the issue of some voxels containing only image or point cloud features. CONet\cite{wang2023openoccupancy} and Occgen\cite{wang2024occgen} use 3D convolutions to apply weighted operations on the image and point cloud feature branches and perform fusion through weighted summation, thus improving the effectiveness and robustness of the fusion results. Unlike the fusion schemes mentioned above, we introduce 3D Gaussian rendering to implement a cross-supervision alignment paradigm, which forms the basis for constructing a more effective fusion module.

\section{Method}
\begin{figure*}[t]
\centering 
\includegraphics[width=1\textwidth]{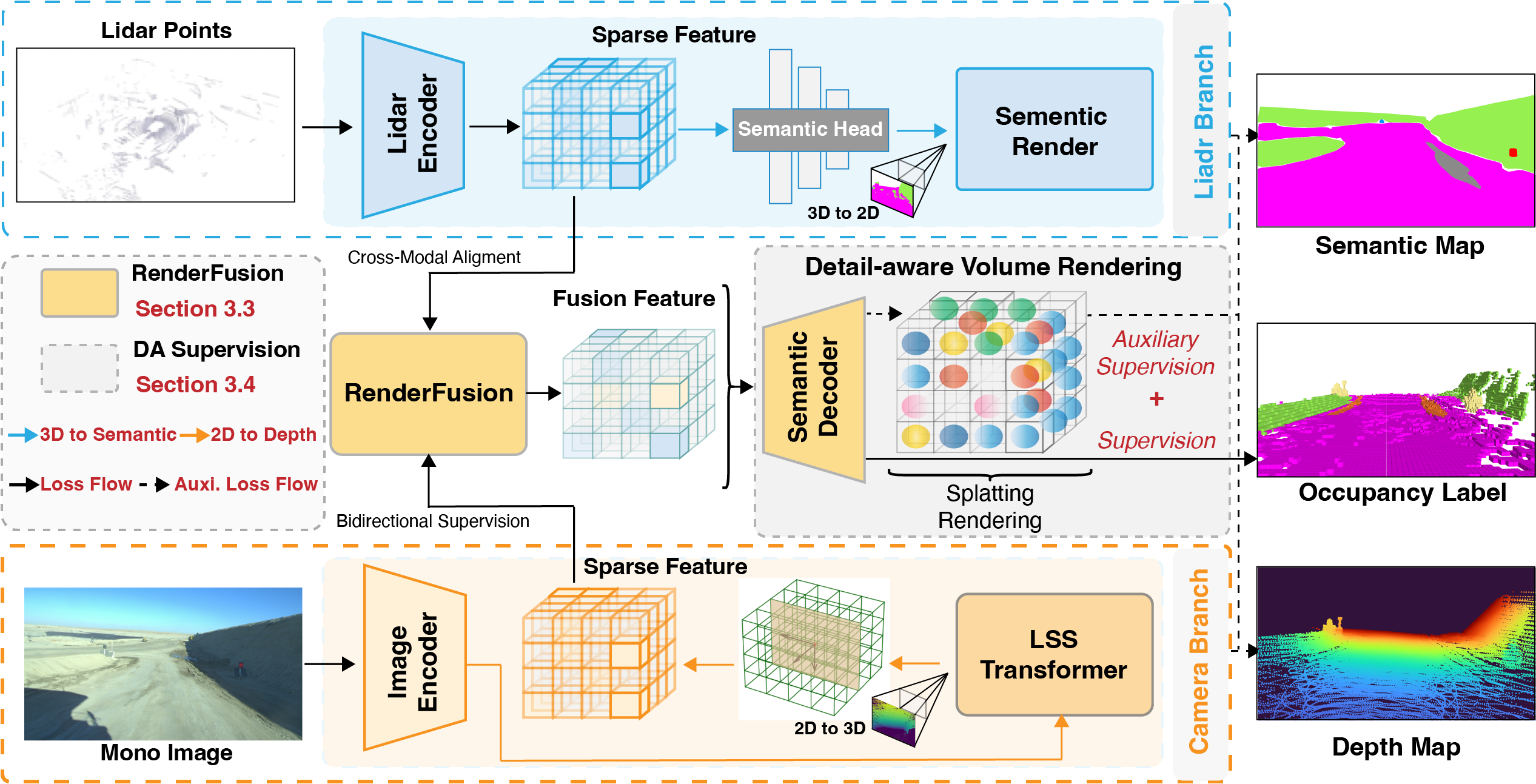}
\caption{\textbf{Framework of our UnsOcc.}
Features from image and LiDAR modalities are extracted, aligned via RenderFusion, and fused. The fused features are used for 3D occupancy prediction, with auxiliary supervision provided by 2D semantic rendering through 3D Gaussian Splatting.
}
\label{fig:framework}
\vspace{-3mm}
\end{figure*}
\subsection{Overview}
The overall architecture of our model is illustrated in Figure \ref{fig:framework}. At its core, we incorporate 3D Gaussian Splatting, a technique that has recently been widely adopted due to its faster rendering speed compared to volume rendering. By representing a scene as a set of 3D Gaussians, each Gaussian distribution $G$ is defined as
\begin{equation}G(X)=e^{-\frac{1}{2}(X-\mu)^T\Sigma^{-1}(X-\mu)},\end{equation}
where $\mu$ and $\Sigma$ denote the mean and 3D covariance matrix, respectively. Through projection, each 3D Gaussian is mapped to a 2D Gaussian $G_{2}$, with its covariance matrix computed as
\begin{equation}\Sigma^\prime=JW\Sigma W^TJ^T,\end{equation}
where $W$ is the view transformation matrix and $J$ the Jacobian matrix. The rendering process then applies alpha-blending to compute pixel colors:
\begin{equation}
c(x)=\sum_{i=1}^Nc^i\alpha^iG_{2}^i(x)\prod_{j=1}^{i-1}\left(1-\alpha^jG_{2}^j(x)\right),
\end{equation}
with $x$ denoting the pixel position, and $c^i$, $\alpha^i$ the color and opacity of the $i$-th Gaussian.

Building on this rendering mechanism, our framework introduces Bidirectional Supervision for Cross-Modal Feature Alignment and Detail-aware Auxiliary Supervision via Gaussian Splatting. The former enhances feature extraction and alignment across modalities through cross-modal supervision, thereby strengthening their fusion. The latter leverages 2D ground truth as an auxiliary supervision signal to supplement non-empty semantic samples, effectively addressing the sparsity of map elements in unstructured scenes.



\subsection{Feature Extraction}
Our framework adopts a dual-branch backbone to process the image and LiDAR modalities in parallel.

For the image branch, we use a ResNet backbone pretrained on ImageNet to capture hierarchical visual patterns. The multi-level features are aggregated by a SECONDFPN neck, which enhances semantic abstraction while preserving spatial details, and outputs unified 128-dimensional representations suitable for downstream tasks.

For the LiDAR branch, the raw point cloud is voxelized into regular 3D grids, with each voxel retaining up to a fixed number of points. A voxel feature encoder based on mean pooling aggregates local geometry, and a sparse convolutional encoder further extracts high-dimensional descriptors that efficiently encode structural and spatial information.

In this way, the image branch contributes rich semantic cues such as textures and object boundaries, while the LiDAR branch provides precise geometric and depth-aware representations. The complementary features are subsequently integrated in the fusion module to enable robust cross-modal reasoning.
\subsection{Bidirectional Supervision for Cross-Modal Feature Alignment.}

To address the fusion challenges arising from the lack of semantic information in the LiDAR branch and depth information in the cameras branch, we propose a bidirectional supervision for cross-modal feature alignment mechanism called RenderFusion as shown in Fig.\ref{fig:framework}. This method introduces semantic supervision to the LiDAR branch and depth supervision to the camera branch through a rendering-based approach.

\noindent\textbf{Depth Rendering for Camera Branch.} 
In the image branch, since the monocular image lacks depth information, we utilize the depth projection from the LiDAR branch as a supervisory signal to supervise the depth prediction in the image branch. In this way, depth information that is geometrically aligned with the LiDAR branch is introduced into the image branch, enhancing its spatial understanding and facilitating cross-modal consistency. 

Specifically, given an input image $I\!\in\! \mathbb{R}^{H\times W\times 3}$, feature $F_I\!\in\! \mathbb{R}^{H\times W\times C}$ is first extracted through a backbone and neck network and then passed through a Depth Net \cite{philion2020lift} to produce a depth distribution $p_d\!\in \!\mathbb{R}^{D\times H\times W}=\Phi_d(F_I)$. $D$ represents the number of depth bins. $F_I$ is lifted into image-depth space by weighted aggregation based on the depth distribution:
\begin{equation}
    F_\text{img-d}(u,v,d)={p_d}(u,v,d)\cdot F_I(u,v).
\end{equation}
By leveraging intrinsic $K$ and extrinsic $E$, the discrete depth coordinates of each pixel $(u,v,d)$ can be mapped to corresponding 3D voxel coordinates without learnable parameter:
\begin{equation}
    F_{\text{vi}}(i,j,k)=\sum_{T(u,v,d)=(i,j,k)} F_\text{img-d}(u,v,d),
\end{equation}
where \( F_\text{vi} \) and \( T \) denote the 3D voxel features of the image and the transformation function that maps image depth coordinates to 3D voxel coordinates, respectively.

The subsequent fusion of voxel-level features is highly dependent on the geometric alignment between the two modalities, whereas the transformation of image features from 2D to 3D at this stage critically relies on the accuracy of depth prediction. To ensure accurate and consistent depth prediction, during training, the point cloud input $P \!\in \!\mathbb{R}^{N \times 4}$ from the LIDAR branch is projected to generate a depth map $Z_L\!\in\!\mathbb{R}^{H\times W}$. By normalizing $Z_{\text{L}}$ and applying one-hot encoding, we obtain the ground truth for the depth distribution $Z_p\!\in \!\mathbb{R}^{D\times H\times W}$:
\begin{equation}
Z_L(u^{\prime}, v^{\prime}) = min(P_{\text{img}}(u, v)),\quad
    P_{\text{img}} = K \cdot [R \mid t] \cdot P^{h},
\end{equation}

where $P^{\text{h}}$ denotes the homogeneous coordinates of a LiDAR point, 
$[R \mid t]$ is the extrinsic matrix, 
$K$ is the intrinsic matrix, 
$P_{\text{img}}\!\in\!\mathbb{R}^{D\times H\times W}$ is the projected 2D image coordinate. $u^{\prime}$ and $v^{\prime}$ are the results of rounding the $u$ and $v$ components, which represent the horizontal and vertical components of the image coordinates, respectively.

\noindent\textbf{Semantic Rendering for Lidar Branch.} 
In the LIDAR branch, we also aim to leverage information from the image branch to improve the alignment between LIDAR and image features. Semantic prediction from LIDAR signals has been widely validated as effective. To facilitate this, we introduce a semantic prediction head that converts 3D LIDAR features into 3D semantic logits. These logits are then projected onto the image plane using 3D Gaussian Splatting and supervised with 2D semantic segmentation maps obtained from the input images.

Concretely, let $P \!\in\! \mathbb{R}^{N \times 4}$ be the input point clouds. The voxelized  $P_{v} \!\in\! \mathbb{R}^{M \times 4}$ contains the mean values of $M$ non-empty voxels. Building upon the voxelized representation, subsequent sparse convolutions hierarchically aggregate local voxel neighborhoods, resulting in sparse LIDAR feature representations $F_L$. Prior to multi-modal fusion, we establish semantic consistency between sparse LIDAR features and image features through an auxiliary learning task. During training, the sparse feature $F_{L} \!\in\! \mathbb{R}^{M \times C}$ is processed by a lightweight segmentation head $\Phi_{seg3d}$ to produce volumetric semantic logits $\mathcal{L}_L \!\in\! \mathbb{R}^{X\times Y\times Z\times K}$. $K$ is the number of 3D occupancy classes. To leverage the information from the image branch as a supervision signal, $\mathcal{L}^3_L$ is then rendered to the 2D image plane via alpha-blending:
\begin{equation}
    \mathcal{L}^2_L = \sum_{i=1}^{N}\mathcal{L}_L\cdot\alpha_i^{\prime}\prod_{j=1}^{i-1}(1-\alpha^{\prime}_i),
\end{equation}
where $N$ represents the number of Gaussians involved in the projection, $\mathcal{L}^2_L$ is the 2D logits obtained from the projection, $\alpha^{\prime}$ is the effective opacity computed from the original opacity and Gaussian density.
At this point, we can supervise the LIDAR voxel features $F_L$ using 2D semantic segmentation $S_I$ that matches the image input.

After pre-fusion optimization, the enhanced LiDAR and image features are fused in 3D voxel space. The LiDAR and projected image features are first spatially aligned, and for each non-empty LiDAR voxel, neighboring image features are retrieved to provide semantic guidance. These neighboring features are then weighted and combined with the LiDAR features to produce the final fused representation.



\subsection{Detail-aware auxiliary supervision via Gaussian splatting}
In unstructured scenes, particularly in large-scale open environments or natural settings, some object classes typically occupy only a small fraction of the space, with the majority of the area remaining empty. Consequently, more than 98\% of the voxels in the 3D occupancy ground truth are empty, resulting in an extremely low density of meaningful semantic occupancy data. This severe sparsity substantially limits the effectiveness of semantic class supervision.
However, in the image space, the proportion of empty classes is significantly reduced. To leverage this advantage, we propose a dense 2D auxiliary supervision mechanism, which employs 3D Gaussian Splatting to exploit semantic information from the image space for 3D occupancy prediction. This strategy substantially increases the number of learning samples for non-empty categories, thereby enhancing the model’s ability to accurately perceive these classes, especially for rare classes.

In particular, the fused features $F_{f}$ is passed through a decoder and a semantic head to output 3D occupancy logits predictions $\mathcal{O}$. Moreover, We employ 3D Gaussian Splatting to project $\mathcal{O}$ onto the camera plane, yielding 2D segmentation predictions $\mathcal{L}^2_{\text{occ}}$ and depth predictions $\mathcal{L}_{d}$\:
\begin{equation}
    \mathcal{L}^2_{occ} = \sum_{i=1}^{N}\mathcal{L}_{occ}\cdot\alpha_i^{\prime}\prod_{j=1}^{i-1}(1-\alpha^{\prime}_i),
\end{equation}
\begin{equation}
    \mathcal{L}^2_{d} = \sum_{i=1}^{N}d_i\cdot\alpha_i^{\prime}\prod_{j=1}^{i-1}(1-\alpha^{\prime}_i),
\end{equation}where $\mathcal{L}_{occ}$ represents the occupancy logits produced by the Semantic Decoder, and $d_i$ denotes the center depth of the i-th Gaussian (the z-coordinate in the camera coordinate system).

Building upon the use of 3D occupancy ground truth for supervision, we also use 2D ground truth, which matches the input image, to supervise the predictions, refining the learning of long tail samples.
\subsection{Loss Function}
\noindent\textbf{Depth Predict Loss.}
The binary cross-entropy (BCE) loss is adopted to supervise the discrete depth distribution:
\begin{equation}
\mathbf{L}_{d} = - \left[ Z_p \log(p_d) + (1 - Z_p) \log(1 - p_d) \right],
\end{equation}
where $p_d$ denotes the predicted depth probability and $Z_p$ represents the corresponding ground-truth label.

\noindent\textbf{Depth Render Loss.}
The SILog loss calculates the loss between the depth predicted $Z_r$ by 3D Gaussian Splatting and the ground truth depth $Z_L$:
\begin{equation}
   \mathbf{L}_{dr}=\frac{1}{N}\sum^{N}_{i=1}|\log(Z_r(i))-\log(Z_L(i))|.
\end{equation}
\noindent\textbf{Segmentation Render Loss.}
Cross-entropy loss is applied to both the 2D logits rendered from LIDAR features $\mathcal{L}^{2}_L$ and those rendered from 3D occupancy predictions $\mathcal{L}^{2}_{occ}$:
\begin{equation}
\mathbf{L}_{2D} = - \log \left( \mathrm{Sigmoid}(\mathcal{L}^{2}_L)_y \right) - \log \left( \mathrm{Sigmoid}(\mathcal{L}^{2}_{occ})_y \right)
\end{equation}
where $y$ represents the ground truth class label.
With the addition of the 3D occupancy loss $\mathbf{L}_{occ}$, the total loss is expressed as follows:
\begin{equation}
\mathbf{L} = \mathbf{L}_{occ}+\mathbf{L}_{d}+\mathbf{L}_{dr}+\mathbf{L}_{2D}.
\end{equation}

\definecolor{color0}{HTML}{000000}  
\definecolor{color1}{HTML}{FF7832}  
\definecolor{color2}{HTML}{A020F0}  
\definecolor{color3}{HTML}{FFC0CB}  
\definecolor{color4}{HTML}{0096F5}  
\definecolor{color5}{HTML}{FFFF00}  
\definecolor{color6}{HTML}{00FFFF}  
\definecolor{color7}{HTML}{FF0000}  
\definecolor{color8}{HTML}{FFF096}  
\definecolor{color9}{HTML}{8B8989}  
\definecolor{color10}{HTML}{96F050} 
\definecolor{color11}{HTML}{FF00FF} 
\definecolor{color12}{HTML}{800000}  
\definecolor{color13}{HTML}{008000}  
\definecolor{color14}{HTML}{000080}  
\definecolor{color15}{HTML}{FFA500}  

\newcommand{\colorboxtext}[2]{
  \tikz[baseline=(char.base)]{
    \node[shape=rectangle, draw, fill=#1, minimum width=0.36cm, minimum height=0.36cm] (char) {};
  } #2
}

{
\renewcommand{\arraystretch}{1.2}  
\begin{table*}[ht]
    \centering
      \setlength{\tabcolsep}{4pt}
    \begin{adjustbox}{width=\textwidth}
    \begin{tabular}{l|c|ccccccccccc|c|c}
     \specialrule{0.5pt}{0pt}{0pt} 
        \hline
        Method & Modality
         &\rotatebox{90}{\colorboxtext{color11}{driveable surface (73.99\%)}} 
         & \rotatebox{90}{\colorboxtext{color10}{terrian (23.70\%)}} 
         & \rotatebox{90}{\colorboxtext{color9}{muddy (0.12\%)}} 
         & \rotatebox{90}{\colorboxtext{color8}{traffic sign (0.41\%)}} 
         & \rotatebox{90}{\colorboxtext{color1}{barrier (0.34\%)}} 
         & \rotatebox{90}{\colorboxtext{color2}{truck (1.04\%)}} 
         & \rotatebox{90}{\colorboxtext{color3}{widebody (0.007\%)}} 
         & \rotatebox{90}{\colorboxtext{color4}{car (0.01\%)}} 
         & \rotatebox{90}{\colorboxtext{color5}{excavator (0.14\%)}} 
         & \rotatebox{90}{\colorboxtext{color6}{machinery (0.22\%)}} 
         & \rotatebox{90}{\colorboxtext{color7}{pedestrian (0.007\%)}} & mIoU & \makecell{mIoU\\(long-tail)}
          \\
        \hline
        
        MonoScene\cite{cao2022monoscene}  & C & 30.11 & 4.82 & 0.10 & 0.62 & 1.91 & 5.09 & 0.00 & 0.12 & 0.00 & 0.41 & 0.00 & 3.92 & 0.92 \\ 

        TPVFormer\cite{huang2023tri} & C & 30.08 & 5.14 & 0.15 & 0.50 & 4.07 & 5.15 & 0.00 & 0.02 & 0.00 & 0.69 & 0.00 & 4.17 & 1.18 \\ 

        Occformer\cite{zhang2023occformer} & C & 30.93 & 8.85 & 11.16 & 6.95 & 5.91 & 8.98 & 0.00 & 3.76 & 0.00 & 6.91 & 0.00 &  4.69  & 4.85\\ 
        CGformer\cite{CGFormer} & C & \underline{\textbf{61.07}} & 31.05 & 5.78 & 16.25 & 10.92 & 22.92 & 0.22 & 4.69 & 8.06 & 11.96 & 0.00  & 15.72 & 8.98\\ 
        L2COcc(C)\cite{wang2025l2cocc} & C & 60.35 & 30.99 & 4.34 & 18.70 & \underline{\textbf{12.07}} & 23.53 & 0.08 & 7.69 & 2.75 & 15.38 & 0.00  & 15.99  & 9.39\\ 
        Co-Occ(L)\cite{pan2024co} & L & 37.26 & 29.72 & 7.72 & 19.47 & 9.93 & 31.4 & 0.00 & 20.92 & 12.81 & 25.98 & 0.00  & 17.75 & 14.25\\ 
        Co-Occ(C\&L)\cite{pan2024co} & C\&L & 37.86 & 29.68 & 8.34 & \underline{\textbf{21.3}} & 9.43 & 32.6 & 0.16 & 19.21 & \underline{\textbf{16.61}} & 28.38 & 0.00  &  18.5 & 15.11\\ 

        \hline
        UnsOcc(Ours) & C\&L & 37.24 & 29.37 & \underline{\textbf{14.70}} & 21.09 & 8.89 & \underline{\textbf{32.17}} & \underline{\textbf{15.55}} & \underline{\textbf{23.1}} & 13.63 & \underline{\textbf{30.55}} & 0.00 & \underline{\textbf{20.57}} & \underline{\textbf{17.74}}\\ 
        
        \hline
         \specialrule{0.5pt}{0pt}{0pt} 
    \end{tabular}
    \end{adjustbox}
    \caption{Evaluation and comparison on open-pit mine test set.}
\label{tab:compare}
\vspace{-6mm}
\end{table*}
}

\section{Results}
\subsection{Datasets}


\noindent\textbf{Open-pit Mine Dataset.} We evaluate our proposed model on a real-world unstructured scene dataset, collected from multiple open-pit mines in China. The sensing platform is equipped with a front-mounted LiDAR, a monocular camera, and an inertial navigation system (INS). Data were gathered under diverse conditions, including daytime, nighttime, rain, snow, and dusty weather, covering both structured roads and unstructured terrains such as muddy or rugged surfaces. In total, the dataset comprises 135 sequences with an average length of 55 frames, featuring complex terrain, loosely distributed objects, and irregular obstacles. Each frame provides synchronized LiDAR point clouds and monocular images.

\begin{figure}[t]
    \centering
    \includegraphics[width=\linewidth]{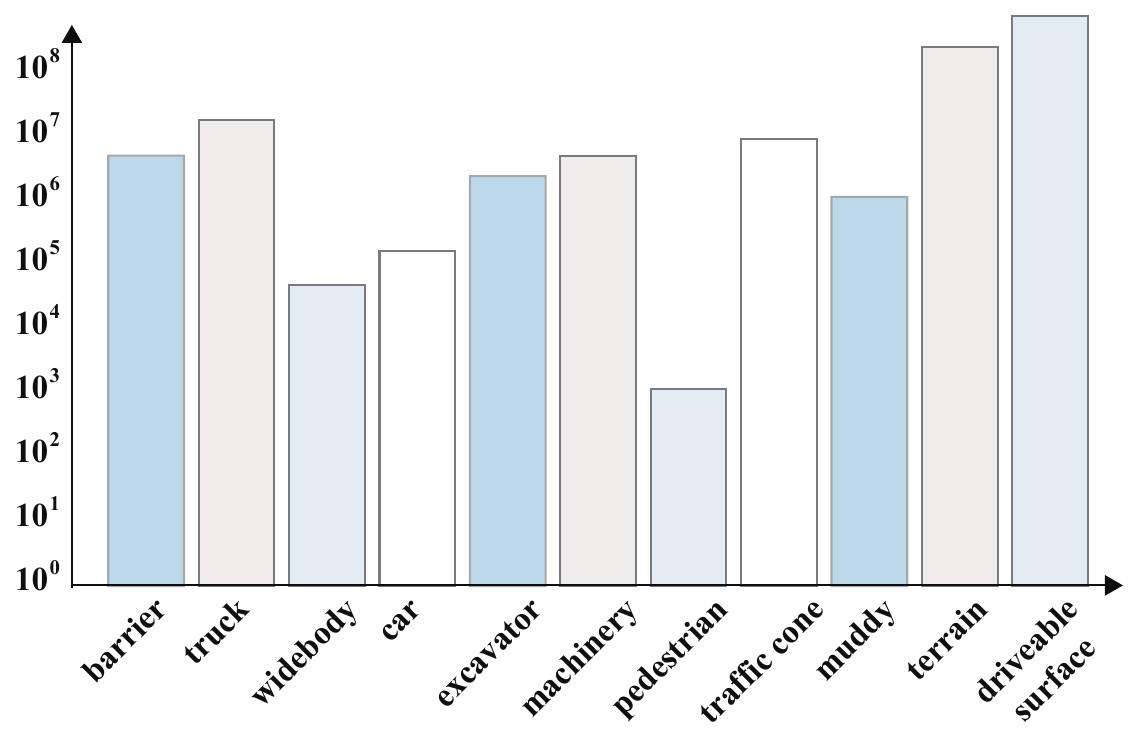}
    \caption{Distribution of semantic classes in the dataset.}
    \label{fig:class_distribution}
    \vspace{-8mm}
\end{figure}

For ground-truth construction, dynamic objects were removed from multi-sweep LiDAR data to generate dense static point cloud maps, which were then voxelized into 3D semantic occupancy labels. The occupancy space is represented as a voxel grid of size [256, 256, 32], with a voxel resolution of [0.3m, 0.3m, 0.3m]. All voxels are categorized into 13 classes, including 11 semantic classes (e.g., driveable surface, terrian, muddy regions, traffic signs, barrier, truck, widebody, car, excavator, machinery and pedestrian), along with an empty and an unknown class. To better illustrate the dataset composition, Fig.~\ref{fig:class_distribution} shows the voxel distribution across these categories. The semantic voxel distribution is highly sparse, where “drivable surface” and “terrain” dominate the majority while other categories occur rarely, leading to a pronounced long-tail distribution. These sparse categories can be considered as long-tail categories. Furthermore, the dataset contains a variety of irregular obstacles and unstructured elements, which increase the difficulty of perception and make it well-suited for evaluating robust 3D occupancy prediction.
Among the 135 sequences, 100 are used for training, 10 for validation, and 25 for testing. Model performance on 3D occupancy prediction is evaluated on the test set using mean Intersection-over-Union (mIoU) and the average IoU over long-tail classes (mIoU(long-tail)).

\noindent\textbf{NuScenes Dataset.} nuScenes\cite{nuscenes} is a large-scale autonomous driving benchmark dataset, consisting of 1,000 driving scenes, of which 700 are used for training, 150 for validation, and 150 for testing. The dataset adopts 3D occupancy labels provided in \cite{wei2023surroundocc}, comprising 17 categories in total, including 16 semantic classes and one empty class. The voxel resolution is set to [0.5 m, 0.5 m, 0.5 m], corresponding to a voxel grid of size [200, 200, 16].

\subsection{Evaluation Metrics}
We evaluate the 3D occupancy prediction performance of our model using mean Intersection-over-Union (mIoU) and Intersection-over-Union (IoU), defined as follows:
\begin{equation}
\mathrm{IoU}=\frac{TP_{\neq c_0}}{TP_{\neq c_0}+FP_{\neq c_0}+FN_{\neq c_0}}
\end{equation}
\begin{equation}
\mathrm{mIoU} = \frac{1}{|\mathcal{C}|-1}\sum_{i\in\mathcal{C},i\neq c_0} \frac{TP_i}{TP_i+FP_i+FN_i}
\end{equation}
where $\mathcal{C}$ denotes the set of all classes, and $c_0$ represents the empty class.
In addition, we also report the mIoU over long-tail categories by excluding \textit{drivable surface} and \textit{terrain} from the semantic classes in the Opne-pit Mine dataset.

\subsection{Implementation details}
Experiments are conducted on 8 NVIDIA L40 GPUs with a batch size of 1 per GPU. Models use ResNet-50 for the Open-pit Mine dataset and ResNet-101 for nuScenes, trained with AdamW (lr=0.0001, weight decay=0.01) for 24 epochs. Baselines follow their official implementations.
\subsection{Main Results}

\begin{figure*}[t]
\centering 
\includegraphics[width=1\textwidth]{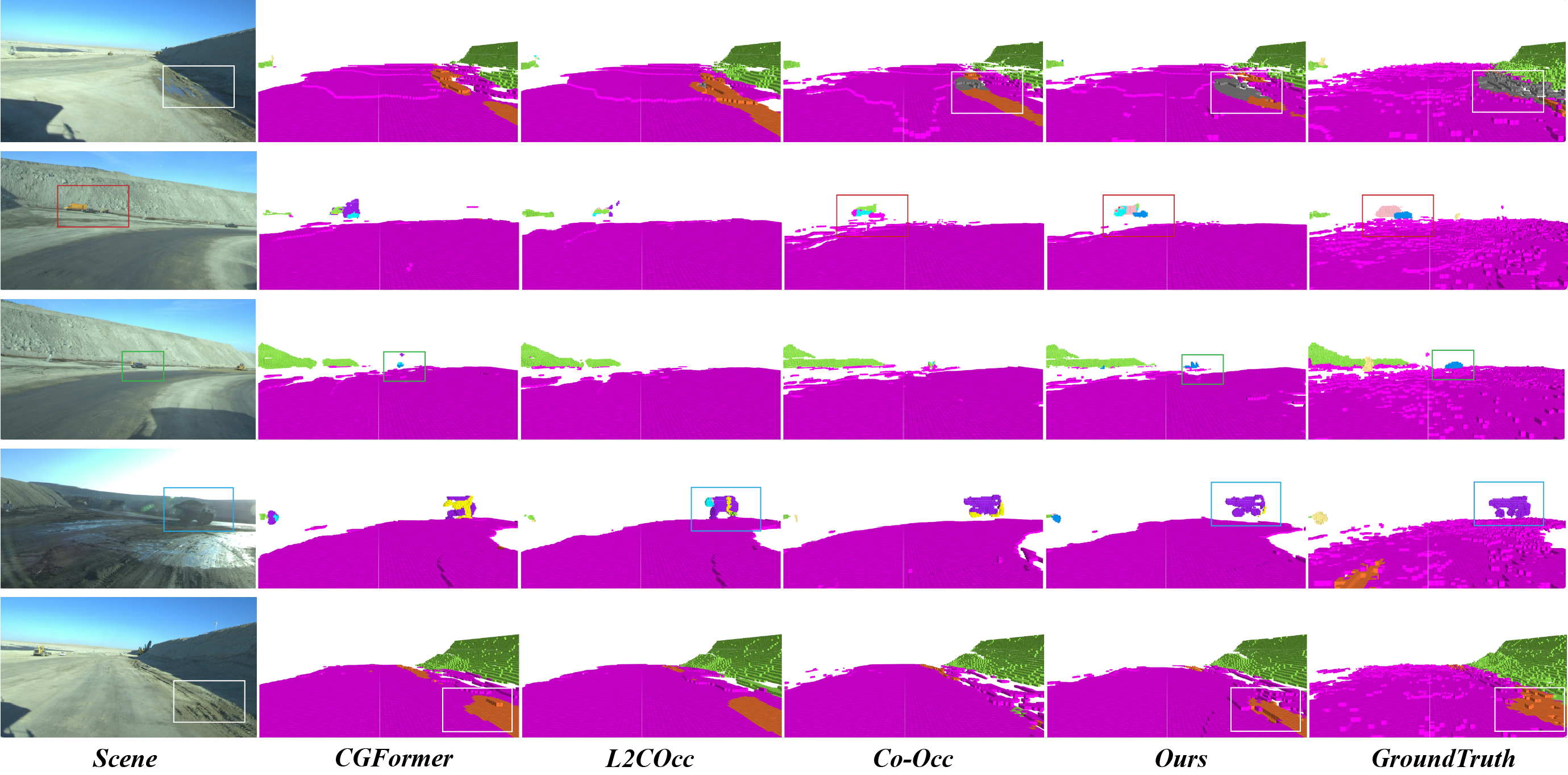}
\caption{
Comparison of the performance of different 3D semantic occupancy prediction methods.
}
\vspace{-2mm}
\label{fig:occ-vis}
\end{figure*}

{
\renewcommand{\arraystretch}{1.2}
\setlength{\tabcolsep}{4pt}
\begin{table*}[ht]
    \centering
    \begin{adjustbox}{width=\textwidth}
    \begin{tabular}{l|c|cccccccccccccccc|c}
     \specialrule{0.5pt}{0pt}{0pt} 
        \hline
        Method & Modality
         &\rotatebox{90}{\colorboxtext{color1}{barrier}}
         & \rotatebox{90}{\colorboxtext{color2}{bicycle}}
         & \rotatebox{90}{\colorboxtext{color3}{bus}}
         & \rotatebox{90}{\colorboxtext{color4}{car}}
         & \rotatebox{90}{\colorboxtext{color5}{const. veh.}}
         & \rotatebox{90}{\colorboxtext{color6}{motorcycle}}
         & \rotatebox{90}{\colorboxtext{color7}{pedestrain}}
         & \rotatebox{90}{\colorboxtext{color8}{traffic cone}}
         & \rotatebox{90}{\colorboxtext{color9}{trailer}}
         & \rotatebox{90}{\colorboxtext{color10}{truck}}
         & \rotatebox{90}{\colorboxtext{color11}{drive. suf.}}
         & \rotatebox{90}{\colorboxtext{color12}{other flat}}
         & \rotatebox{90}{\colorboxtext{color13}{sidewalk}}
         & \rotatebox{90}{\colorboxtext{color14}{terrian}}
         & \rotatebox{90}{\colorboxtext{color15}{manmade}}
         & \rotatebox{90}{\colorboxtext{color0}{vegetation}}
         & mIoU
          \\
        \hline
        
        MonoScene\cite{cao2022monoscene} & C & 4.0 & 0.4 & 8.0 & 8.0 & 2.9 & 0.3 & 1.2 & 0.7 & 4.0 & 4.4 & 27.7 & 5.2 & 15.1 & 11.3 & 9.0 & 14.9 & 7.3\\ 

        BEVFormer\cite{bevformer} & C & 14.2 & 6.5 & 23.4 & 28.2 & 8.6 & 10.7 & 6.4 & 4.0 & 11.2 & 17.7 & 37.2 & 18.0 & 22.8 & 22.1 & 13.8 & 22.2 & 16.7\\ 

        SurroundOcc\cite{wei2023surroundocc} & C & 20.5 & 11.6 & 28.1 & 30.8 & 10.7 & 15.1 & 14.0 & 12.0 & 14.3 & 22.2 & 37.2 & 23.7 & 24.4 & 22.7 & 14.8 & 21.8 & 20.3\\ 
        OccFormer\cite{zhang2023occformer} & C & 21.1 & 11.3 & 28.2 & 30.3 & 10.6 & 15.7 & 14.4 & 11.2 & 14.0 & 22.6 & 37.3 & 22.4 & 24.9 & 23.5 & 15.2 & 21.1 & 20.1\\ 
        C-CONet\cite{wang2023openoccupancy} & L & 18.6 & 10.0 & 26.4 &27.4 & 8.6 & 15.7 & 13.3 & 9.7 & 10.9 & 20.2 & 33.0 & 20.7 & 21.4 & 21.8 & 14.7 & 21.3 & 18.4\\ 
        FB-OCc\cite{li2023fb} & C & 20.6 & 11.3 & 26.9 & 29.8 & 10.4 & 13.6 & 13.7 & 11.4 & 11.5 & 20.6 & 38.2 & 21.5 & 24.6 & 22.7 & 14.8 & 21.6 & 19.6\\ 
        RenderOcc\cite{pan2024renderocc} & C & 19.7 & 11.2 & 28.1 & 28.2 & 9.8 & 14.7 & 11.8 & 11.9 & 13.1 & 20.1 & 33.2 & 21.3 & 22.6 & 22.3 & 15.3 & 20.9 & 19.0\\ 
        LMSCNet\cite{roldao2020lmscnet} & L & 13.1 & 4.5 & 14.7 & 22.1 & 12.6 & 4.2 & 7.2 & 7.1 & 12.2 & 11.5 & 26.3 & 14.3 & 21.1 & 15.2 & 18.5 & 34.2  & 14.9\\
        L-CONet\cite{wang2023openoccupancy} & L & 19.2 & 4.0 & 15.1 & 26.9 & 6.2 & 3.8 & 6.8 & 6.0 & 14.1 & 13.1 & 39.7 & 19.1 & 24.0 & 23.9 & 25.1 & 35.7 & 17.7\\
        M-CONet\cite{wang2023openoccupancy} & C\&L & 24.8 & 13.0 & 31.6 & 34.8 & 14.6 & 18.0 & 20.0 & 14.7 & 20.0 & 26.6 & 39.2 & 22.8 & 26.1 & 26.0 & 26.0 & 37.1 & 24.7\\
        Co-Occ\cite{pan2024co} & C\&L & 28.1 & 16.1 & 34.0 & \underline{\textbf{37.7}} & 17.0 & 21.6 & 20.8 & 15.9 & 21.9 & 28.7 & 42.3 & 25.4 & 29.1 & \underline{\textbf{28.6}} & 28.2 & 38.0 & 27.1\\
        \hline
        UnsOcc(Ours) & C\&L & \underline{\textbf{29.8}} & \underline{\textbf{18.2}} & \underline{\textbf{34.5}} & 37.2 & \underline{\textbf{18.2}} & \underline{\textbf{22.1}} & \underline{\textbf{21.6}} & \underline{\textbf{22.7}} & \underline{\textbf{22.0}} & \underline{\textbf{29.9}} & \underline{\textbf{42.6}} & \underline{\textbf{26.1}} & \underline{\textbf{29.2}} & 28.4 & \underline{\textbf{28.6}} & \underline{\textbf{38.1}} & \underline{\textbf{28.1}}\\ 
        \hline
         \specialrule{0.5pt}{0pt}{0pt} 
    \end{tabular}
    \end{adjustbox}
    \caption{Evaluation and comparison experiments on nuScenes-Occupancy validation set.}
\label{tab:compare}
\vspace{-4mm}
\end{table*}
}

\noindent\textbf{Results on Open-pit Mine Dataset.} Experiments are conducted on the Open-pit Mine Dataset, comparing the proposed approach with various methods with different input modalities, including image-only\cite{cao2022monoscene, huang2023tri, zhang2023occformer, CGFormer, wang2025l2cocc}, LiDAR-only\cite{pan2024co}, and LiDAR-image fusion\cite{pan2024co} methods. Table \ref{tab:compare} shows the per-class IoU and overall mIoU. Early image-only methods \cite{cao2022monoscene, huang2023tri, zhang2023occformer} perform poorly. Compared to recent image-only approaches \cite{CGFormer, wang2025l2cocc}, our method improves mIoU by 30.6\% and 28.6\%.
Against recent LiDAR-only methods \cite{pan2024co}, our method achieves 15.9\% higher mIoU, showing the benefit of multi-modal fusion. Compared to LiDAR-image fusion methods, our approach achieves an 11.19\% mIoU gain, with notable improvements on rare classes. It is worth noting that our method achieves better performance on the muddy class compared to all other baselines.
The lack of LiDAR returns in muddy regions significantly limits the effectiveness of LiDAR-based methods. Even advanced fusion strategies, such as Co-Occ\cite{pan2024co}, fail to yield notable gains in these challenging areas. In contrast, our RenderFusion leverages image-guided rendering supervision to compensate for the LiDAR deficiency, leading to notable improvements and demonstrating the effectiveness and robustness of our approach. Moreover, our method achieves a significant improvement in mIoU (long-tail) over all the above methods, demonstrating its superiority in handling long-tail categories. Figure~\ref{fig:occ-vis} presents the qualitative comparisons between our method and existing approaches.\\
\noindent\textbf{Results on nuScenes Dataset.} To further assess the generalizability of our model, we conduct supplementary experiments on the nuScenes. Our approach achieves higher mIoU compared to several previous state-of-the-art methods, demonstrating its capability to capture both coarse and fine-grained scene structures. Our method shows clear advantages on challenging objects such as traffic cones and pedestrians, indicating stronger sensitivity to fine details. These observations collectively support the robustness of our model and its applicability across different datasets and scenarios.

\subsection{Ablation Study}
\label{sec:result}
\begin{table}[ht]
    \centering
    \setlength{\tabcolsep}{2pt}

    \begin{adjustbox}{width=0.5\textwidth}
    \begin{tabular}{c|c|c|c|c|c}

     \specialrule{1pt}{0pt}{0pt} 

        GSRefinement & RenderFusion & Resolution & modality & mIoU& \makecell{mIoU\\(long-tail)}\\\hline
        \ding{55} & \ding{55} & 384×1280 & C\&L & 18.50 & 15.11\\
        \ding{51} & \ding{55} & 384×1280 & C\&L & 19.63 & 16.56\\
        \ding{55} & \ding{51} & 384×1280 & C\&L & 19.08 & 15.86\\
        \ding{51} & \ding{51} & 384×1280 & C\&L & 20.57 & 17.74\\
        \hline
         \specialrule{0.5pt}{0pt}{0pt} 
    \end{tabular}
    \end{adjustbox}
    \caption{Ablation study on architectural components.}
\label{tab:ablation}
\vspace{-6mm}
\end{table}

\noindent\textbf{Effect of the GSRefinement.} 
As shown in Table \ref{tab:ablation}, incorporating the Detail-aware module into the baseline leads to a 1.13 improvement in mIoU. Furthermore, the mIoU(long-tail classes) improves by 1.45. This demonstrates that the Detail-aware component effectively enhances the perception capability. The introduction of 2D supervision signals significantly improves semantic prediction performance, especially for rare classes.

\noindent\textbf{Effect of the RenderFusion.} 
We further evaluate the effectiveness of the proposed RenderFusion module. As shown in Table \ref{tab:ablation}, introducing RenderFusion leads to an additional improvement of 0.55 in mIoU and 0.75 in rare class mIoU. This demonstrates that the proposed multi-modal feature alignment strategy further enhances the quality of fused features and improves recognition performance on both common and rare semantic categories.

\noindent\textbf{Combination of GSRefinement and RenderFusion.}
When both GSRefinement and RenderFusion are introduced, the mIoU and mIoU (long-tail) further improve. The result shows that introducing RenderFusion on top of GSRefinement brings more gains. Conversely, introducing GSRefinement on top of RenderFusion also yields additional improvements.

\subsection{Deployment}
We deployed our proposed method on different vehicle types at multiple mining sites, with vehicles equipped with monocular and multi-view cameras. The deployment results show that the fusion strategy provides significant gains with monocular input and remains effective under multi-view settings, consistent with the aforementioned experimental results, further validating the robustness and practicality of our approach. Additional qualitative results are available in the supplementary video.
\section{Conclusion}
\label{sec:conclusion}
This paper proposes a multimodal 3D occupancy prediction framework that integrates RenderFusion with GSRefinement, leveraging complementary image and LiDAR information to enhance scene understanding. The framework introduces 3D Gaussian Splatting to enable cross-modal supervision between branches, improving feature alignment before voxel-level fusion. It further incorporates semantically rich 2D signals for complementary supervision, boosting occupancy prediction performance. Experiments on the open-pit mine and nuScenes datasets validate the effectiveness of the proposed method.

\bibliographystyle{IEEEtran}
\bibliography{example}

@inproceedings{lang2019pointpillars,
  title     = {Pointpillars: Fast encoders for object detection from point clouds},
  author    = {Lang, Alex H and Vora, Sourabh and Caesar, Holger and Zhou, Lubing and Yang, Jiong and Beijbom, Oscar},
  booktitle = {Proceedings of the IEEE/CVF Conference on Computer Vision and Pattern Recognition (CVPR)},
  pages     = {12697--12705},
  year      = {2019},
  organization={IEEE}
}

@inproceedings{liu2023bevfusion,
  title     = {Bevfusion: Multi-task multi-sensor fusion with unified bird's-eye view representation},
  author    = {Liu, Zhijian and Tang, Haotian and Amini, Alexander and Yang, Xinyu and Mao, Huizi and Rus, Daniela L and Han, Song},
  booktitle = {2023 IEEE International Conference on Robotics and Automation (ICRA)},
  pages     = {2774--2781},
  year      = {2023},
  organization={IEEE}
}

@inproceedings{cao2022monoscene,
  title     = {Monoscene: Monocular 3d semantic scene completion},
  author    = {Cao, Anh-Quan and De Charette, Raoul},
  booktitle = {Proceedings of the IEEE/CVF Conference on Computer Vision and Pattern Recognition (CVPR)},
  pages     = {3991--4001},
  year      = {2022},
  organization={IEEE}
}

@inproceedings{wei2023surroundocc,
  title     = {Surroundocc: Multi-camera 3d occupancy prediction for autonomous driving},
  author    = {Wei, Yi and Zhao, Linqing and Zheng, Wenzhao and Zhu, Zheng and Zhou, Jie and Lu, Jiwen},
  booktitle = {Proceedings of the IEEE/CVF International Conference on Computer Vision (ICCV)},
  pages     = {21729--21740},
  year      = {2023},
  organization={IEEE}
}

@inproceedings{zhang2023occformer,
  title     = {Occformer: Dual-path transformer for vision-based 3d semantic occupancy prediction},
  author    = {Zhang, Yunpeng and Zhu, Zheng and Du, Dalong},
  booktitle = {Proceedings of the IEEE/CVF International Conference on Computer Vision (ICCV)},
  pages     = {9433--9443},
  year      = {2023},
  organization={IEEE}
}

@article{pan2024co,
  title     = {Co-occ: Coupling explicit feature fusion with volume rendering regularization for multi-modal 3d semantic occupancy prediction},
  author    = {Pan, Jingyi and Wang, Zipeng and Wang, Lin},
  journal   = {IEEE Robotics and Automation Letters},
  year      = {2024},
  publisher = {IEEE}
}

@inproceedings{wang2023openoccupancy,
  title     = {Openoccupancy: A large scale benchmark for surrounding semantic occupancy perception},
  author    = {Wang, Xiaofeng and Zhu, Zheng and Xu, Wenbo and Zhang, Yunpeng and Wei, Yi and Chi, Xu and Ye, Yun and Du, Dalong and Lu, Jiwen and Wang, Xingang},
  booktitle = {Proceedings of the IEEE/CVF International Conference on Computer Vision (ICCV)},
  pages     = {17850--17859},
  year      = {2023},
  organization={IEEE}
}

@article{ming2024occfusion,
  title   = {Occfusion: A straightforward and effective multi-sensor fusion framework for 3d occupancy prediction},
  author  = {Ming, Zhenxing and Stephany Berrio, Julie and Shan, Mao and Worrall, Stewart},
  journal = {arXiv preprint arXiv:2403.00000},
  year    = {2024}
}

@inproceedings{chen2020novel,
  title     = {A novel calibration method between a camera and a 3D LiDAR with infrared images},
  author    = {Chen, Shoubin and Liu, Jingbin and Liang, Xinlian and Zhang, Shuming and Hyypp{\"a}, Juha and Chen, Ruizhi},
  booktitle = {2020 IEEE International Conference on Robotics and Automation (ICRA)},
  pages     = {4963--4969},
  year      = {2020},
  organization={IEEE}
}

@inproceedings{huang2024gaussianformer,
  title={Gaussianformer: Scene as gaussians for vision-based 3d semantic occupancy prediction},
  author={Huang, Yuanhui and Zheng, Wenzhao and Zhang, Yunpeng and Zhou, Jie},
  booktitle={European Conference on Computer Vision},
  pages={376--393},
  year={2024},
  organization={Springer}
 }

@inproceedings{shi2024occupancy,
  title     = {Occupancy as set of points},
  author    = {Shi, Yiang and Cheng, Tianheng and Zhang, Qian and Liu, Wenyu and Wang, Xinggang},
  booktitle = {Proceedings of the European Conference on Computer Vision (ECCV)},
  pages     = {72--87},
  year      = {2024},
  organization={Springer}
}

@inproceedings{li2023voxformer,
  title     = {Voxformer: Sparse voxel transformer for camera-based 3d semantic scene completion},
  author    = {Li, Yiming and Yu, Zhiding and Choy, Christopher and Xiao, Chaowei and Alvarez, Jose M and Fidler, Sanja and Feng, Chen and Anandkumar, Anima},
  booktitle = {Proceedings of the IEEE/CVF Conference on Computer Vision and Pattern Recognition (CVPR)},
  pages     = {9087--9098},
  year      = {2023},
  organization={IEEE}
}

@inproceedings{wang2024occgen,
  title     = {Occgen: Generative multi-modal 3d occupancy prediction for autonomous driving},
  author    = {Wang, Guoqing and Wang, Zhongdao and Tang, Pin and Zheng, Jilai and Ren, Xiangxuan and Feng, Bailan and Ma, Chao},
  booktitle = {Proceedings of the European Conference on Computer Vision (ECCV)},
  pages     = {95--112},
  year      = {2024},
  organization={Springer}
}

@article{kerbl20233d,
  title   = {3D Gaussian Splatting for Real-Time Radiance Field Rendering},
  author  = {Kerbl, Bernhard and Kopanas, Georgios and Leimk{\"u}hler, Thomas and Drettakis, George},
  journal = {ACM Transactions on Graphics},
  volume  = {42},
  number  = {4},
  pages   = {139:1--139:14},
  year    = {2023}
}

@inproceedings{philion2020lift,
  title     = {Lift, splat, shoot: Encoding images from arbitrary camera rigs by implicitly unprojecting to 3d},
  author    = {Philion, Jonah and Fidler, Sanja},
  booktitle = {Proceedings of the European Conference on Computer Vision (ECCV)},
  pages     = {194--210},
  year      = {2020},
  organization={Springer}
}

@article{mildenhall2021nerf,
  title     = {NeRF: Representing scenes as neural radiance fields for view synthesis},
  author    = {Mildenhall, Ben and Srinivasan, Pratul P and Tancik, Matthew and Barron, Jonathan T and Ramamoorthi, Ravi and Ng, Ren},
  journal   = {Communications of the ACM},
  volume    = {65},
  number    = {1},
  pages     = {99--106},
  year      = {2021},
  publisher = {ACM}
}

@article{zhang2020nerf++,
  title   = {NeRF++: Analyzing and Improving Neural Radiance Fields},
  author  = {Zhang, Kai and Riegler, Gernot and Snavely, Noah and Koltun, Vladlen},
  journal = {arXiv preprint arXiv:2010.07492},
  year    = {2020}
}

@inproceedings{barron2021mip,
  title     = {Mip-NeRF: A multiscale representation for anti-aliasing neural radiance fields},
  author    = {Barron, Jonathan T and Mildenhall, Ben and Tancik, Matthew and Hedman, Peter and Martin-Brualla, Ricardo and Srinivasan, Pratul P},
  booktitle = {Proceedings of the IEEE/CVF International Conference on Computer Vision (ICCV)},
  pages     = {5855--5864},
  year      = {2021},
  organization={IEEE}
}

@inproceedings{pan2024renderocc,
  title     = {RenderOcc: Vision-centric 3D occupancy prediction with 2D rendering supervision},
  author    = {Pan, Mingjie and Liu, Jiaming and Zhang, Renrui and Huang, Peixiang and Li, Xiaoqi and Xie, Hongwei and Wang, Bing and Liu, Li and Zhang, Shanghang},
  booktitle = {2024 IEEE International Conference on Robotics and Automation (ICRA)},
  pages     = {12404--12411},
  year      = {2024},
  organization={IEEE}
}

@article{zhang2023occnerf,
  title   = {OccNeRF: Self-supervised multi-camera occupancy prediction with neural radiance fields},
  author  = {Zhang, Chubin and Yan, Juncheng and Wei, Yi and Li, Jiaxin and Liu, Li and Tang, Yansong and Duan, Yueqi and Lu, Jiwen},
  journal = {CoRR},
  year    = {2023}
}

@article{jiang2024gausstr,
  title   = {GaussTR: Foundation Model-Aligned Gaussian Transformer for Self-Supervised 3D Spatial Understanding},
  author  = {Jiang, Haoyi and Liu, Liu and Cheng, Tianheng and Wang, Xinjie and Lin, Tianwei and Su, Zhizhong and Liu, Wenyu and Wang, Xinggang},
  journal = {arXiv preprint arXiv:2412.13193},
  year    = {2024}
}

@article{gan2024gaussianocc,
  title   = {GaussianOcc: Fully Self-supervised and Efficient 3D Occupancy Estimation with Gaussian Splatting},
  author  = {Gan, Wanshui and Liu, Fang and Xu, Hongbin and Mo, Ningkai and Yokoya, Naoto},
  journal = {arXiv preprint arXiv:2408.11447},
  year    = {2024}
}

@article{chambon2025gaussrender,
  title   = {GaussRender: Learning 3D Occupancy with Gaussian Rendering},
  author  = {Chambon, Lo{\~A}{\NG}ck and Zablocki, Eloi and Boulch, Alexandre and Chen, Micka{\~A}Ŧl and Cord, Matthieu},
  journal = {arXiv preprint arXiv:2502.05040},
  year    = {2025}
}

@inproceedings{chabot2025gaussianbev,
  title     = {GaussianBEV: 3D Gaussian Representation Meets Perception Models for BEV Segmentation},
  author    = {Chabot, Florian and Granger, Nicolas and Lapouge, Guillaume},
  booktitle = {2025 IEEE/CVF Winter Conference on Applications of Computer Vision (WACV)},
  pages     = {2250--2259},
  year      = {2025},
  organization={IEEE}
}

@article{miao2023occdepth,
  title   = {OccDepth: A Depth-Aware Method for 3D Semantic Scene Completion},
  author  = {Miao, Ruihang and Liu, Weizhou and Chen, Mingrui and Gong, Zheng and Xu, Weixin and Hu, Chen and Zhou, Shuchang},
  journal = {arXiv preprint arXiv:2302.13540},
  year    = {2023}
}

@article{tian2023occ3d,
  title   = {Occ3D: A Large-Scale 3D Occupancy Prediction Benchmark for Autonomous Driving},
  author  = {Tian, Xiaoyu and Jiang, Tao and Yun, Longfei and Mao, Yucheng and Yang, Huitong and Wang, Yue and Wang, Yilun and Zhao, Hang},
  journal = {Advances in Neural Information Processing Systems},
  volume  = {36},
  pages   = {64318--64330},
  year    = {2023}
}

@inproceedings{tong2023scene,
  title     = {Scene as Occupancy},
  author    = {Tong, Wenwen and Sima, Chonghao and Wang, Tai and Chen, Li and Wu, Silei and Deng, Hanming and Gu, Yi and Lu, Lewei and Luo, Ping and Lin, Dahua and others},
  booktitle = {Proceedings of the IEEE/CVF International Conference on Computer Vision (ICCV)},
  pages     = {8406--8415},
  year      = {2023},
  organization={IEEE}
}

@inproceedings{huang2023tri,
  title     = {Tri-Perspective View for Vision-Based 3D Semantic Occupancy Prediction},
  author    = {Huang, Yuanhui and Zheng, Wenzhao and Zhang, Yunpeng and Zhou, Jie and Lu, Jiwen},
  booktitle = {Proceedings of the IEEE/CVF Conference on Computer Vision and Pattern Recognition (CVPR)},
  pages     = {9223--9232},
  year      = {2023},
  organization={IEEE}
}

@inproceedings{ob3,
  title     = {MVX-Net: Multimodal VoxelNet for 3D Object Detection},
  author    = {Sindagi, Vishwanath A and Zhou, Yin and Tuzel, Oncel},
  booktitle = {2019 International Conference on Robotics and Automation (ICRA)},
  pages     = {7276--7282},
  year      = {2019},
  organization={IEEE}
}

@inproceedings{seg2,
  title     = {MSEG3D: Multi-Modal 3D Semantic Segmentation for Autonomous Driving},
  author    = {Li, Jiale and Dai, Hang and Han, Hao and Ding, Yong},
  booktitle = {Proceedings of the IEEE/CVF Conference on Computer Vision and Pattern Recognition (CVPR)},
  pages     = {21694--21704},
  year      = {2023},
  organization={IEEE}
}

@article{wang2025l2cocc,
  title   = {L2COcc: Lightweight Camera-Centric Semantic Scene Completion via Distillation of LiDAR Model},
  author  = {Wang, Ruoyu and Ma, Yukai and Yao, Yi and Tao, Sheng and Li, Haoang and Zhu, Zongzhi and Liu, Yong and Zuo, Xingxing},
  journal = {arXiv preprint arXiv:2503.12369},
  year    = {2025}
}

@inproceedings{CGFormer,
  title     = {Context and Geometry Aware Voxel Transformer for Semantic Scene Completion},
  author    = {Yu, Zhu and Zhang, Runmin and Ying, Jiacheng and Yu, Junchen and Hu, Xiaohai and Luo, Lun and Cao, Si-Yuan and Shen, Hui-liang},
  booktitle = {Advances in Neural Information Processing Systems (NeurIPS)},
  pages     = {1531--1555},
  volume    = {37},
  year      = {2024}
}

@inproceedings{seg1,
  title     = {UniSeg: A Unified Multi-Modal LiDAR Segmentation Network and the OpenPCSeg Codebase},
  author    = {Liu, Youquan and Chen, Runnan and Li, Xin and Kong, Lingdong and Yang, Yuchen and Xia, Zhaoyang and Bai, Yeqi and Zhu, Xinge and Ma, Yuexin and Li, Yikang and others},
  booktitle = {Proceedings of the IEEE/CVF International Conference on Computer Vision (ICCV)},
  pages     = {21662--21673},
  year      = {2023},
  organization={IEEE}
}

@inproceedings{ob4,
  title     = {UniTR: A Unified and Efficient Multi-Modal Transformer for Bird's-Eye-View Representation},
  author    = {Wang, Haiyang and Tang, Hao and Shi, Shaoshuai and Li, Aoxue and Li, Zhenguo and Schiele, Bernt and Wang, Liwei},
  booktitle = {Proceedings of the IEEE/CVF International Conference on Computer Vision (ICCV)},
  pages     = {6792--6802},
  year      = {2023},
  organization={IEEE}
}

@inproceedings{nuscenes,
  title={nuscenes: A multimodal dataset for autonomous driving},
  author={Caesar, Holger and Bankiti, Varun and Lang, Alex H and Vora, Sourabh and Liong, Venice Erin and Xu, Qiang and Krishnan, Anush and Pan, Yu and Baldan, Giancarlo and Beijbom, Oscar},
  booktitle={Proceedings of the IEEE/CVF conference on computer vision and pattern recognition},
  pages={11621--11631},
  year={2020}
}

@article{bevformer,
  title={Bevformer: learning bird's-eye-view representation from lidar-camera via spatiotemporal transformers},
  author={Li, Zhiqi and Wang, Wenhai and Li, Hongyang and Xie, Enze and Sima, Chonghao and Lu, Tong and Yu, Qiao and Dai, Jifeng},
  journal={IEEE Transactions on Pattern Analysis and Machine Intelligence},
  year={2024},
  publisher={IEEE}
}

@article{li2023fb,
  title={Fb-occ: 3d occupancy prediction based on forward-backward view transformation},
  author={Li, Zhiqi and Yu, Zhiding and Austin, David and Fang, Mingsheng and Lan, Shiyi and Kautz, Jan and Alvarez, Jose M},
  journal={arXiv preprint arXiv:2307.01492},
  year={2023}
}

@inproceedings{roldao2020lmscnet,
  title={Lmscnet: Lightweight multiscale 3d semantic completion},
  author={Roldao, Luis and De Charette, Raoul and Verroust-Blondet, Anne},
  booktitle={2020 International Conference on 3D Vision (3DV)},
  pages={111--119},
  year={2020},
  organization={IEEE}
}
\end{document}